# A Novel Index-Based Multidimensional Data Organization Model that Enhances the Predictability of the Machine Learning Algorithms

Mahbubur Rahman

Department of Computer Science, North American University,
Stafford, Texas, USA

## Abstract

*Learning from the multidimensional data has been an interesting concept in the field of machine learning. However, such learning can be difficult, complex, expensive because of expensive data processing, manipulations as the number of dimension increases. As a result, we have introduced an ordered index-based data organization model as the ordered data set provides easy and efficient access than the unordered one and finally, such organization can improve the learning. The ordering maps the multidimensional dataset in the reduced space and ensures that the information associated with the learning can be retrieved back and forth efficiently. We have found that such multidimensional data storage can enhance the predictability for both the unsupervised and supervised machine learning algorithms.*

## 1. Introduction

Searching, classifying, predicting the multidimensional data have been the most interesting applications of today's machine learning algorithms [1],[2]. As the number of dimensions, size of data increase, so does the overall complexity of pre-processing, searching, classifying or predicting such data sets using machine learning algorithms [3]. To overcome such complexity, we have introduced a data organization model that can enhance overall learning outcome while keeping the necessary information associated with the learning in a reduced dimensional space.

The multidimensional data requires high dimensional data storage, access models. There are several high dimensional data organization models as SR-tree [4], R-tree [5], kd-tree [6]. These are tree-based data organization models that follow the complexity of tree while inserting, deleting multidimensional data. One of the recent patents from Google [7] has explained the methods of searching the multidimensional dataset by mapping the data to the nodes in the multidimensional hyperspace. However, we are interested to design a data organization model for the multidimensional data in a reduced space that has efficient data access, manipulation schema along with the stored information that can be used in the learning outcome of the machine learning algorithms.

To design such multidimensional data organization model, we have introduced a multidimensional index-based data organization model that can be used in the machine learning algorithm. This data organization model is ordered on the Euclidean norm and stores the original index position along with the Euclidean norm of the datasets. This ordering allows searching in logarithmic time and index mapping to the original database in constant time. As a result, the overall data organization model provides a reduced space along with the original index-Euclidean norm pair that can be reused in the machine learning computation enhancing the overall learning outcome.

The following sections and subsections explain the overall data organization model, its applications on the supervised, unsupervised machine learning models, implementations, results and analysis.

## 2. DATA ORGANIZATION MODEL

The data organization model is a two-dimensional map (e.g. reduced space) of the original multidimensional dataset (e.g. Fig. 1) that keeps the information (e.g. original index position and Euclidean norm) by the increasing order of the Euclidean norm of the datasets. As a result, this information can be accessed and reused efficiently. As for example, the Euclidean norm can be reused in the Euclidean distance [8], cosine similarity [9] calculation etc. Additionally, such ordered data organization model can provide the searching spaces for a point of interest that can enhance finding its nearest neighbours, clusters etc. Thus, the ordered index-based database provides important information that can enhance the predictability of the machine learning algorithms. We have provided further details in the following subsections.

| Index | Dim 1 | Dim 2 | Dim 3 | ... | Dim n |
|-------|-------|-------|-------|-----|-------|
| 0 | $a_1$ | $a_2$ | $a_3$ | ... | $a_n$ |
| 1 | $b_1$ | $b_2$ | $b_3$ | ... | $b_n$ |
| 2 | $c_1$ | $c_2$ | $c_3$ | ... | $c_n$ |
| ... | ... | ... | ... | ... | ... |
| m | $z_1$ | $z_2$ | $z_3$ | ... | $z_n$ |

Original N Dimensional Datasets

| Index | Original Index | Euclidean Norm |
|-------|----------------|----------------|
| 0 | 2 | $\sqrt{c_1^2 + c_2^2 + \cdots + c_n^2}$ |
| 1 | 0 | $\sqrt{a_1^2 + a_2^2 + \cdots + a_n^2}$ |
| 2 | m | $\sqrt{z_1^2 + z_2^2 + \cdots + z_n^2}$ |
| ... | ... | ... |
| m | 1 | $\sqrt{b_1^2 + b_2^2 + \cdots + b_n^2}$ |

Ordered index-based Database

Figure 1. Original multidimensional datasets and ordered index-based data organization model.

### 2.1. Ordered Index-Based Database

The ordered database is organized from the original data source. It is ordered by the Euclidean norm and stores both the original index position and the Euclidean norm. As a result, the Euclidean norm can be mapped to the original index position of the data directly. This index and Euclidean norm can be stored as a pair to a hash table that forms the ordered database (e.g. Fig. 1).

The Euclidean norm of an n dimensional vector (e.g. x) has the following formula:

$$\text{Euclidean\_norm(x)} = \sqrt{x1^2 + x2^2 + \cdots + xn^2} \qquad (1)$$

The Euclidean distance between the two multidimensional vector (e.g. x, y) has the following formula:

Euclidean_distance(x,y) =

$$\sqrt{(x1-y1)^2 + (x2-y2)^2 + \cdots + (xn-yn)^2} \quad (2)$$

$$=\sqrt{x_1^2 + \dots + x_n^2 + y_1^2 + \dots + y_n^2 - 2\sum_{i=1}^{n} x_i . y_i^T} \quad \text{[e.g. expanding } (x-y)^2 \text{ formula] (3)}$$

$$=\sqrt{Euclidean\_norm(x)^2 + Euclidean\_norm(y)^2 - 2\sum_{i=1}^{n} x_i . y_i^T}$$

[e.g. from (1)] (4)

Hence, the unknown part of formula (4) is the dot product of the two vectors. The rest is available from the Euclidean norm of the database. This can be reused in calculating the cosine similarity of the two multidimensional vectors as suggested by the following cosine formula:

$$\frac{\sum_{i=1}^{n} x_i y_i}{Euclidean\_norm(x) \quad Euclidean\_norm(y)} \quad (5)$$

The denominator part of (5) is also available from the database. Additionally, such organization of the multidimensional data sets provides the nearest neighbour searching space for a dataset. It means that the nearest neighbours for a data can exist within a minimum distance between two immediate neighbours of the dataset as it is explained in Figure 2.

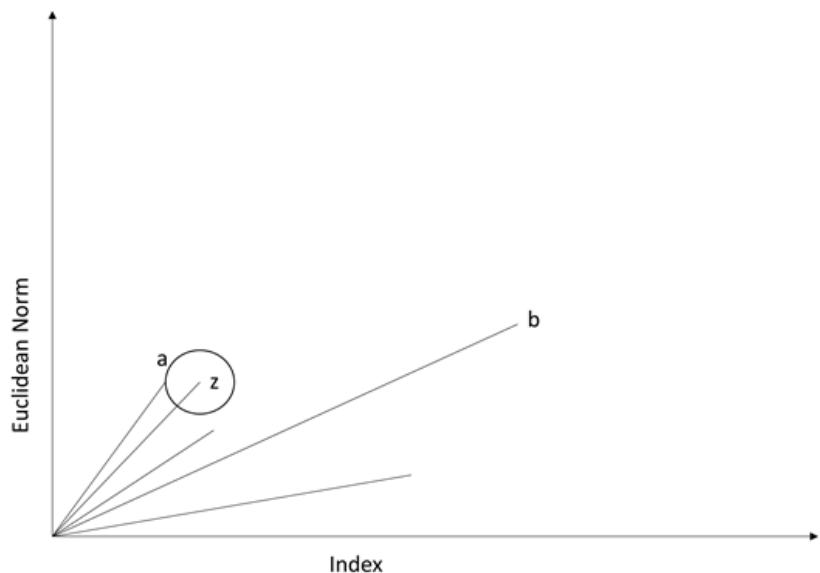

Figure 2. Index vs Euclidean Norm of the ordered database. The nearest neighbours of z exist in the circle having the radius of the minimum distance of its immediate neighbour (e.g. a).

We find from the Fig. 2 is that the two immediate neighbours are one index up and down (e.g. left and right of z in Fig. 2) respectively from the index position of the data of interest in the ordered database. Next, the minimum Euclidean distance between these two immediate neighbours provides the searching space (e.g. radius of the circle in Fig. 2) for the nearest neighbours of the dataset of interest. The index positions of the dataset in the ordered database can be retrieved in logarithmic time by the binary search and then their original index position in the database in constant time (e.g. Fig. 1). As a result, the ordered database can provide the searching space information that can be used in the supervised and unsupervised machine learning algorithms. The overall searching space for a point of interest along with the reuse of the above computations has been listed as Algorithm 1: Nearest Neighbour Searching Algorithm (NNSA) below.

---

**ALGORITHM 1:** Nearest Neighbor Searching Algorithm (NNSA)

---

Function NNSA (x, ordered_db, original_db):
**Input:** A point (x), ordered database (ordered_db), original database (original_db).
**Output:** Two immediate neighboring index (e.g. ui, di) of x.
index_pos_x, euclidean_norm_x = ordered_db[x]
index_up = index_pos_x+1
index_down = index_pos_x-1
xu, euclidean_norm_up = ordered_db[index_up]
xd, euclidean_norm_down = ordered_db[index_down]
x1 = original_db[xu]
x2 = original_db[xd]

$$d1 = \sqrt{euclidean\_norm\_up^2 + euclidean\_norm\_up^2 + \cdots - 2\sum_{i=1}^{n} x_i x_{1i}^T}$$

$$d2 = \sqrt{euclidean\_norm\_down^2 + euclidean\_norm\_down^2 + \cdots - 2\sum_{i=1}^{n} x_i x_{2i}^T}$$

d = min (d1, d2)
d_up = euclidean_norm_x - d
d_down = euclidean_norm_x+d
ui = ordered_db[d_down]
di = ordered_db[d_up]
**return** ui, di

## 2.2. An Organized Data Model Based Supervised Learning

The supervised learning model requires the labelled data sets for learning and predicting the outcome. This labelling information is used during the training steps. The model executes different types of computations (e.g searching, comparing etc.) at the training phase. We have explained how the ordered index-based database can enhance such computations as well as learning the outcome in this subsection.

We have used the K nearest neighbour (KNN) supervised learning model [10] as this is one of the most fundamental supervised learning algorithms. In KNN algorithm, the neighbours for a new point of interest in the dataset are the K closest instances. To locate such K nearest neighbours, the algorithm must first calculate the Euclidean distance between each record of the dataset and the point of interest. Finally, it must sort all the datasets by their distances to the point of interest to get the K required instances. There are several attempts to optimize the KNN using a multi-step query processing strategy [8], GPU computing [11]. However, we are interested to optimize the learning outcome of the KNN using the NNSA.

The NNSA can provide the KNNs for a point of interest while searching around the point. The searching space can be increased by the same minimum amount of the radius of the searching space if the required number of nearest neighbours is not found within the initial searching space. Additionally, the ordered database can be reused again to find the KNNs for a new point of interest. Thus, this model helps avoiding repetitive expensive sort, computation operations. The overall KNN algorithm using the NNSA has been listed as Algorithm 2: KNN Algorithm using NNSA below.

---

**ALGORITHM 2**: KNN Algorithm using NNSA

---

Function K_NNSA (x, ordered_db, original_db, k, neighbor_list):

**Input:** A point (x), ordered database (ordered_db), original database (original_db), number of neighbors (k) of x, list of nearest neighbors (neighbor_list) of x.

**Output:** List of nearest neighbors (neighbor_list) of x.

ui, di = NNSA (x, ordered_db, original_db)
neighbor_counter = 0
neighbor_list = []
u0 = ui
d0 = di
**while** neighbor_counter <= k **do**
    **if** u0 – k <= 0 **then**
        u0 = u0 – k
    **end if**
    **if** d0 + k < length(original_db) **then**
        d0 = d0 + k
    **end if**
    up = u0
    down = d0
    **while** up <= down **do**
        index_pos_up = ordered_db[up]
        label_pos_up = original_db[index_pos_up]
        **if** label_pos_up == x[label_index] **then**
            neighbor_counter = neighbor_counter + 1

        neighbor_list.
            add(original_db[index_pos_up])
        **end if**
        index_pos_down = ordered_db[down]
        label_pos_down = original_db[index_pos_down]
        **if** label_pos_down == x[label_index] **then**
            neighbor_counter = neighbor_counter + 1
            neighbor_list.
            add(original_db[index_pos_up])
        **end if**
        up = up + 1
        down = down -1
        **if** length(neighbor_list) == k **then**
            **return** neighbor_list
    **end while**
**end while**
**return** neighbor_list

### 2.3. An Organized Data Model based Unsupervised Learning

The unsupervised learning algorithm does not require labelled datasets [12], rather can learn from the unlabelled datasets. The existing K-means clustering algorithms calculate the Euclidean distance among the centroids of the clusters and the points until the convergence [13], [14]. As a result, the overall computation involved here are repetitive and thus, expensive. However, we have organized the data such that this repetitive calculation can be reduced from multiple times to single one. Additionally, the NNSA can also determine the minimum no. of clusters and finally, this can be used to determine the optimum no. of clusters.

The ordered database can provide the minimum number of clusters by selecting the mid index point (e.g. centroid) of the ordered database and then determining its searching space (e.g. Fig. 3) by the NNSA. Next, this step can be repeated for the rest of the datasets above and below the searching space of the mid index point until it covers the ordered database. Finally, this minimum no. of clusters can be reduced to the optimum no. of clusters by applying agglomerative [15], divisive [16], hierarchical [17] clustering algorithms.

The optimum number of clusters can be determined where the number of clusters and within cluster sum of squares (WCSS) [18] begins to level off (e.g. elbow method). The WCSS is defined as the sum of the squared distance between each member of the cluster (e.g., x) and its centroid (e.g., c) and has the following formula:

$$\sum_{i=1}^{n}(xi - ci)^2 \qquad\qquad (6)$$

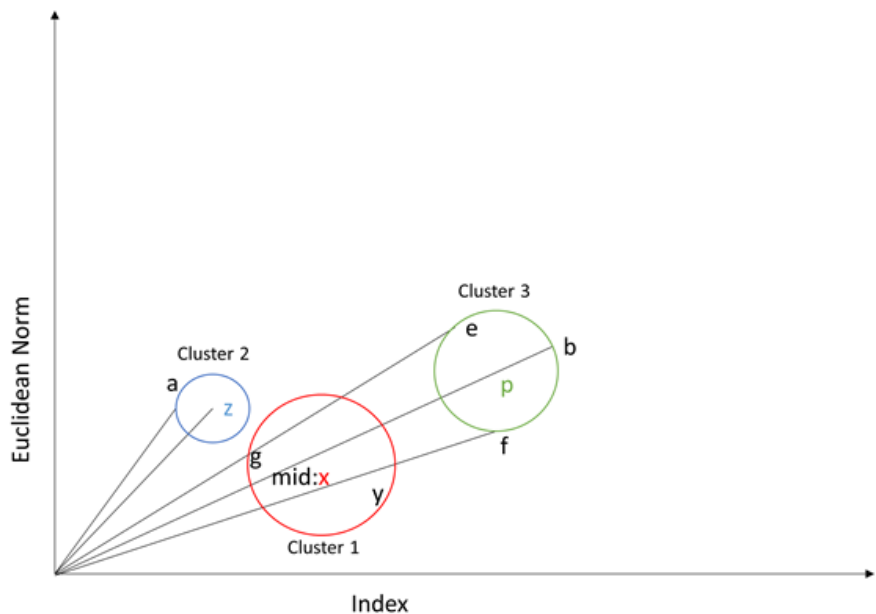

Figure 3. Cluster formation by the ordered index-based database. Cluster 1 starts from the mid indexed vector x (e.g. mid: x in the figure). Similar vectors belong to different clusters depending on the Euclidean distance between the centroid and the cluster

Our proposed ordered database can provide the Euclidean norms to the formula (6) and thus, can enhance the overall computation. The overall K-means algorithm using the NNSA has been listed as Algorithm 3: K-means Algorithm using NNSA below.

**ALGORITHM 3**: K-Means Algorithm using NNSA

Function K-Means_NNSA(x, ordered_db, original_db, centroid_list):
**Input:** Midpoint (x), ordered database (ordered_db), original database (original_db), list of centroids (centroid_list).
**Output:** List of centroids (centroid_list).
up, down = NNSA (x, ordered_db, original_db)
centroid_list = []
centroid_list.add(original_db[x])
up = up / 2
n_points = down – up
max_size = length(original_db)
**while** up > 0 **do**
      l, r = NNSA (up, ordered_db, original_db)
      l = l – (n_points/2)
      **if** l < 0 **then**
            l = 0
      **end if**
      r = r + (n_points/2)
      **if** r >= down-1 **then**
            r = down-1
            add the point at the up index along with l, r to the centroid list
            centroid_list.add(original_db[up])
             break
      **end if**
      up = up / 2
**end while**
down = down + (n_points/2)
**while** down < max_size **do**
      l, r = NNSA (down, ordered_db, original_db)
      l = l – (n_points/2)
      **if** l < 0 **then**
            l = 0
      **end if**
      r = r + (n_points/2)
      **if** r >= max_size **then**
            r = max_size - 1
            centroid_list.add(original_db[down])
             break
      **end if**
      down = down + n_points / 2
**end while**
**return** centroid_list

## 3. IMPLEMENTATIONS

All the algorithms above have been implemented in python version 3 [19]. The ordered database of the NNSA is a python dictionary of the pair of original index position and Euclidean norm of the multidimensional dataset. As result, this original index position can be mapped to the position of the dataset in the original database directly. We have used the existing python libraries to sort the dictionary by the Euclidean norm. The above three algorithms are three different python subroutines where the KNN and K-Means subroutines both interact with the NNSA subroutine. Finally, all the subroutines along with the databases have been incorporated into a python class file.

## 4. RESULTS

We have introduced an index-based data organizational model for the multidimensional data in a reduced space to improve the predictability of the learning algorithms. The reduced space is a two-dimensional space (e.g. a pair of original index and Euclidean norm) for the multidimensional dataset. The organization of this reduced space requires the dataset to be sorted by the Euclidean norm. Hence, there are some computations involved in the organization of the data model. The computations include calculating the Euclidean norm, sorting by that norm and finally, inserting as a pair of original index and Euclidean norm of the dataset into a dictionary. We have defined the time required for such computations as the data model organization time and analysed the effect of increasing the no. of dimensions, sizes of the datasets (e.g. Fig. 4, 5 respectively).

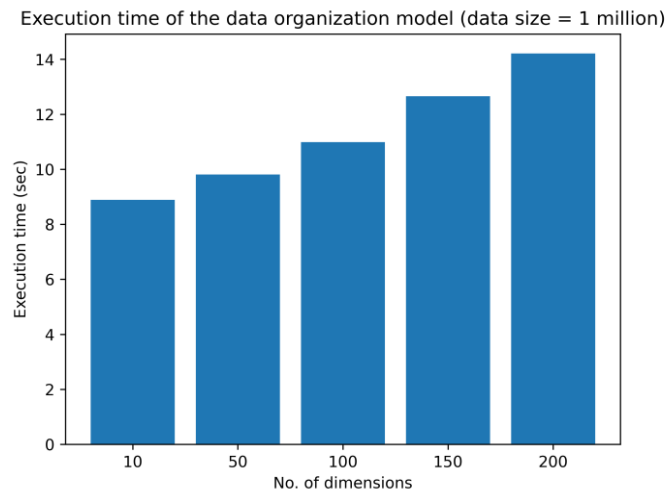

Figure 4. No. of dimensions vs Execution time of the data organization model for a fixed size data (e.g. 1 million).

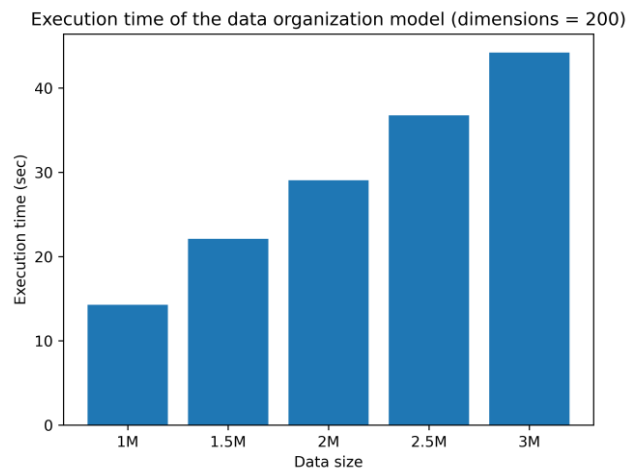

Figure 5. Data size vs execution time of the model for a fixed dimension size (e.g. 200).

We have executed the NNSA on an eight-core intel i7 2.60 GHz, 16 GB main memory PC and received the above performance. It is evident that the NNSA algorithm requires more computing resources for large size multidimensional datasets and can be further improved while incorporating the computing power of the graphical processor unit (e.g. GPU) [20].

Next, we have compared our model with the existing models from the scikit-learn package [21]. We have used the iris dataset [22] to compare between the models that use the NNSA and that don't. We have found that the KNN model that uses the NNSA show better accuracy irrespective to the no. of neighbours than that doesn't (e.g. Fig. 6). As we have already explained in the previous sections that the NNSA provides the searching space of the nearest neighbours for the point of interest enhancing the overall accuracy of the model. The higher accuracy of the KNN with the NNSA requires some little extra time as well (e.g. Fig. 7).

Next, we have tested the above dataset in the K-means algorithms that use the NNSA and don't respectively. We found that the minimum no. of clusters suggested by the K-means with NNSA are eight whereas it is four as suggested by the traditional K-means algorithm (e.g. Fig. 8). The NNSA can determine the minimum no. of clusters by providing the minimum searching space for a point of interest and thus, can help to predict the optimum no. of clusters by forming single clusters following agglomerative, hierarchical approaches [23].

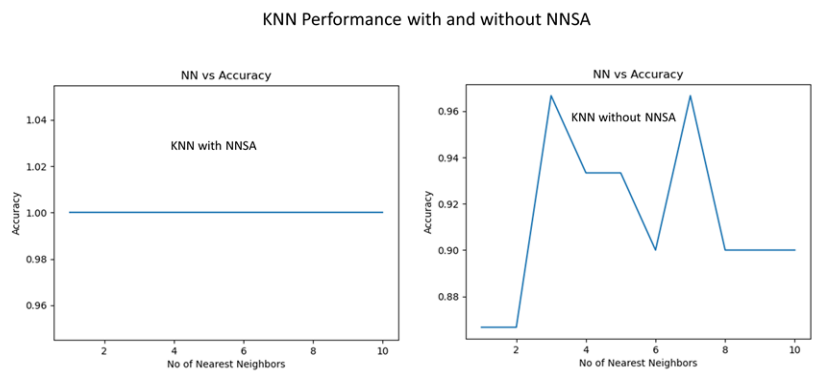

Figure 6. Model performance with and without the NNSA. The KNN with the NNSA shows constant accuracy over the no. of nearest neighbours.

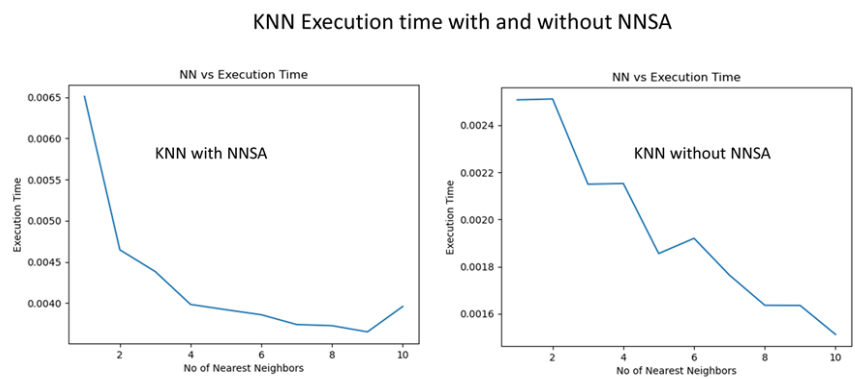

Figure 7. KNN with and without NNSA versus Execution Time (s). The higher accuracy of the KNN with NNSA is responsible for some extra time

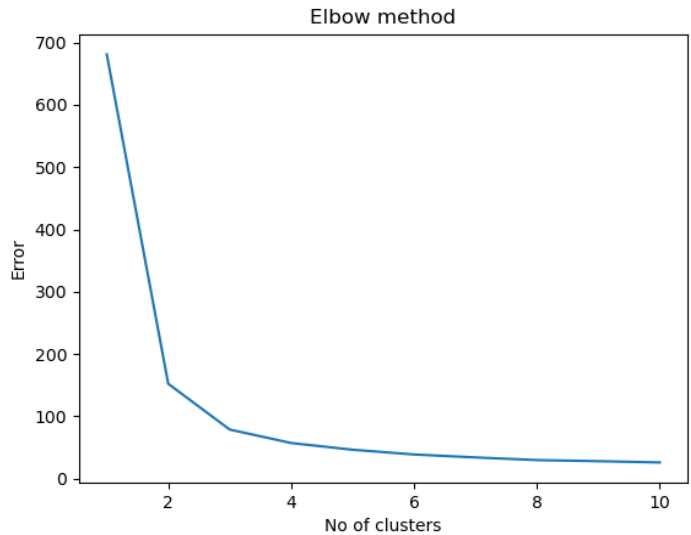

Figure 8. Error vs no. of clusters to determine optimum no. of clusters required for the iris dataset.

# 5. CONCLUSIONS

We have introduced an index-based data organization model for the multidimensional dataset and explained how this organization can enhance the predictability of both the supervised and unsupervised machine learning algorithms. The analysis of the multidimensional data increases as the no. of dimension increases [24]. We have explained that this complexity of analysis can be reduced by organizing the dataset in a reduced space. Our approach is simple, easy to understand and can be integrated with the existing machine learning algorithms. We have used the Euclidean norm as a metric of the data sorting and next, this has been reused in the calculations involved in the distance-based learning algorithms. Similar way, other metrics that have repeating usages in the learning algorithms can be applicable in this data organization model to further improve the predictability, data accessibility and overall computing efficiency of such algorithms.

## ACKNOWLEDGEMENTS

The authors would like to thank the department of computer science at the North American University for supporting the research.

## REFERENCES

[1]   Pat Langley and Herbert A. Simon, "Applications of Machine Learning and Rule Induction", Communications of the ACM November 1995/Vol. 38, No. 11, 1995.

[2]   MI Jordan, TM Mitchell, "Machine learning: Trends, perspectives and prospects", Science 2015.

[3]   G. J. Grevera and A. Meystel, "Searching in a multidimensional space", Proceedings. 5th IEEE International Symposium on Intelligent Control Philadelphia, PA, USA, 1990, pp. 700-705 vol.2, 1990.

[4]   Norio Catyama and Shin'ichi Satoh, " The SR-Tree: An Index Structure For High-Dimensional Nearest Neighbor Queries", *ACM*, 1997.

[5]   Antonin Guttman, "R-Trees: A Dynamic Index Structure For Spatial Searching", *ACM*, 1984.

[6]   Jo-Mei Chang, King-Sun Fu, "Extended K-d Tree Database Organization: A Dynamic Multiattribute Clustering Method", *IEEE Transactions on Software Engineering*, 1981.

[7]   EG Sirer, NL Caruso, B Wong, R Escriva, "System and methods for mapping and searching objects in multidimensional space", US Patent 9,317,536, 2016 - Google Patents

[8]   Thomas Seidl, Hans-Peter Kriegel, "Optimal Multi-Step k-Nearest Neighbor Search" , Proceedings of the 1998 ACM SIGMOD.

[9]   Peipei Xia, Li Zhang, Fanzhang Li, "Learning similarity with cosine similarity ensemble", Information Sciences,Volume 307,2015,Pages 39-52, ISSN 0020-0255.

[10]  D. Sculley, "Web-Scale K-Means Clustering", 2010, Proceedings of the 19th international conference on World wide web, 2010, Pages 1177–1178.

[11]  V. Garcia, E. Debreuve and M. Barlaud, "Fast k nearest neighbor search using GPU", IEEE Computer Society Conference on Computer Vision and Pattern Recognition Workshops, Anchorage, AK, 2008, pp. 1-6.

[12]  K Wagstaff, C Cardie, S Rogers, S Schrödl, "Constrained K-means Clustering with Background Knowledge", Proceedings of the Eighteenth International Conference on Machine Learning, 2001, p. 577–584.

[13]  S. Na, L. Xumin and G. Yong, "Research on k-means Clustering Algorithm: An Improved k-means Clustering Algorithm," 2010 Third International Symposium on Intelligent Information Technology and Security Informatics, Jinggangshan, 2010, pp. 63-67, doi: 10.1109/IITSI.2010.74.

[14]  D. Sculley, "Web-Scale K-Means Clustering", 2010, Proceedings of the 19th international conference on World wide web, 2010, Pages 1177–1178.

[15]   Athman Bouguettaya, Qi Yu, Xumin Liu, Xiangmin Zhou, Andy Song, "Efficient agglomerative hierarchical clustering", Expert Systems with Applications, 2015, Volume 42, Issue 5.

[16]  T Xiong, S Wang, A Mayers, E Monga, "DHCC: Divisive hierarchical clustering of categorical data", Data mining and knowledge discovery, 2011.

[17]  Jianxin Wang, Min Li, Jianer Chen, and Yi Pan, "A Fast Hierarchical Clustering Algorithm for Functional Modules Discovery in Protein Interaction Networks", IEEE/ACM Transactions on computational biology and bioinformatics, 2010.

[18]  M. N. Vrahatis, B. Boutsinas, P. Alevizos, G. Pavlides,"The New k-Windows Algorithm for Improving the k-Means Clustering Algorithm", Mathematical and Computer Modelling, journal of complexity 18, 375–391 (2002).

[19]  Mark Lutz, Learning Python, Third Edition, 2008, Published by O'Reilly Media, Inc.

[20]  N Govindaraju, J Gray, R Kumar, Dinesh Manocha, "GPUTeraSort: High Performance Graphics Co-processor Sorting for Large Database Management", Microsoft Technical Report MSR TR-2005-183, 2006.

[21]  Scikit-Learn: Machine learning in Python.

[22]  R. A. Fisher, Sc.D., F.R.S., "The use of multiple measurements in taxonomic problems", 1936, Annals of eugenics, Wiley Online Library.

[23]  Mark Ming-Tso Chiang, Boris Mirkin, "Intelligent Choice of the Number of Clusters in K-Means Clustering: An Experimental Study with Different Cluster Spreads", Journal of Classification 27 (2009).

[24]  P Indyk, R Motwani," Approximate nearest neighbors: towards removing the curse of dimensionality", Proceedings of the thirtieth annual ACM, 1991.

## AUTHORS

**Mahbubur Rahman** completed his masters from the Texas Tech University in 2010 and PhD from the University of Houston in 2015 in computer science. His research interest is in machine learning, data mining, algorithm optimization. He developed a multidimensional searching algorithm during his master's at the Texas Tech University. He also developed and enhanced computational modeling using machine learning algorithm and HPC environment at the University of Houston during his PhD. He developed a clustering-based prediction algorithm during his postdoctoral research at the University of Texas Medical Branch at Galveston (UTMB).

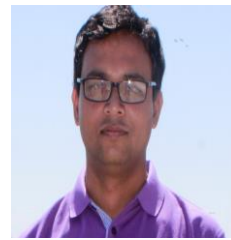

He also developed and managed two scientific websites:
https://cdssim.chem.ttu.edu/
https://dynamic-proteome.utmb.edu

Linkedin profile:
https://www.linkedin.com/in/mahbubur-rahman-26544a27

Github link: https://www.github.com/shawonmr

Google scholar: https://scholar.google.com/citations?user=w7x71U4AAAAJ&hl=en